\documentclass[runningheads]{llncs}

\usepackage{iciapabbrv}

\usepackage{graphicx}
\usepackage{booktabs}

\usepackage{siunitx}

\usepackage{xcolor}
\usepackage{colortbl}
\usepackage{amsmath}

\usepackage[accsupp]{axessibility}  

\usepackage{hyperref}

\usepackage{orcidlink}

\begin{document}

\title{Benchmarking GANs, Diffusion Models, and Flow Matching for T1w-to-T2w MRI Translation} 

\titlerunning{Benchmark on T1w-to-T2w MRI Translation}

\author{Andrea Moschetto\inst{1} \and
Lemuel Puglisi\inst{1}\orcidlink{0009-0003-5661-9873} \and
Alec Sargood\inst{3} \and Pierluigi Dell'Acqua\inst{2} \and Francesco Guarnera\inst{1} \orcidlink{0000-0002-7703-3367} \and Sebastiano Battiato\inst{1}\orcidlink{0000-0001-6127-2470} \and Daniele Ravì\inst{2} \orcidlink{0000-0003-0372-2677}}

\authorrunning{A.~Moschetto et al.}

\institute{Università degli Studi di Catania, Catania CT 95124, ITA\\ \email{andrea.moschetto@studium.unict.it} \and
Università degli Studi di Messina, Messina, ME 98122, ITA \and
University College London, London LDN, WC1E 6BT, UK\\
}

\maketitle

\begin{abstract}
Magnetic Resonance Imaging (MRI) enables the acquisition of multiple image contrasts, such as T1-weighted (T1w) and T2-weighted (T2w) scans, each offering distinct diagnostic insights. However, acquiring all desired modalities increases scan time and cost, motivating research into computational methods for cross-modal synthesis. To address this, recent approaches aim to synthesize missing MRI contrasts from those already acquired, reducing acquisition time while preserving diagnostic quality. Image-to-image (I2I) translation provides a promising framework for this task. In this paper, we present a comprehensive benchmark of generative models—specifically, Generative Adversarial Networks (GANs), diffusion models, and flow matching (FM) techniques—for T1w-to-T2w 2D MRI I2I translation. All frameworks are implemented with comparable settings and evaluated on three publicly available MRI datasets of healthy adults. Our quantitative and qualitative analyses show that the GAN-based Pix2Pix model outperforms diffusion and FM-based methods in terms of structural fidelity, image quality, and computational efficiency. Consistent with existing literature, these results suggest that flow-based models are prone to overfitting on small datasets and simpler tasks, and may require more data to match or surpass GAN performance. These findings offer practical guidance for deploying I2I translation techniques in real-world MRI workflows and highlight promising directions for future research in cross-modal medical image synthesis. Code and models are publicly available at \url{https://github.com/AndreaMoschetto/medical-I2I-benchmark}
\end{abstract}

\section{Introduction}
\label{sec:intro}

Magnetic Resonance Imaging (MRI) is one of the most powerful and widely used medical imaging techniques.  A unique strength of MRI lies in its ability to generate multiple image contrasts — such as T1-weighted (T1w), T2-weighted (T2w), and FLAIR images — each emphasizing different tissue characteristics. For example, T1w images are useful for highlighting fat, while T2w images are more effective at detecting fluid. As a result, multiple scan types are often required for a single patient, which can increase both the duration and cost of the MRI examination. To address these challenges, there is increasing interest in methods that can reduce the number of scans needed without sacrificing diagnostic value. One promising approach is image-to-image (I2I) translation, where computational models estimate missing image contrasts from those already acquired, for example, synthesizing a T2w image starting from a T1w one. By synthesizing additional modalities from existing scans, I2I translation has the potential to accelerate MRI workflows, reduce patient burden, and lower healthcare costs. Furthermore, I2I techniques can help address situations where certain contrasts are missing due to time and cost constraints or incomplete acquisitions, ensuring that clinicians and downstream models still have access to the necessary imaging information for accurate diagnosis and analysis. \newline

Recent advances in generative artificial intelligence have significantly enhanced the generation of synthetic MRI contrasts and have emerged as powerful tools for I2I translation. These methods offer a data-driven approach to synthesizing missing MRI modalities from available ones. Among them, Generative Adversarial Networks (GANs)—such as Pix2Pix~\cite{isola2017image}—have been widely adopted for their ability to produce high-resolution and realistic outputs by learning mappings between paired image domains in an adversarial manner. More recently, diffusion models~\cite{ho2020denoising} have demonstrated superior performance in generating high-fidelity images by iteratively refining noise into structured outputs. Another promising direction is Flow Matching (FM)~\cite{lipman2022flow}, which models transformations between distributions through learned continuous dynamics.\newline 

Despite significant advancements in generative modeling for MRI contrast translation, a comprehensive and fair comparison of recent methods is still lacking. In this paper, we benchmark state-of-the-art generative models for synthesizing T2w 2D axial slices from corresponding T1w slices within a unified experimental framework. To our knowledge, this is the first direct comparison of GANs, diffusion models, and FMs for this task, all implemented with the same U-Net backbone~\cite{ronneberger2015u}. We present both quantitative and qualitative evaluations highlighting the strengths and limitations of each approach.\newline

Our study aims to identify the model that achieves the best image fidelity, structural consistency, and computational efficiency simultaneously, providing clear guidance for deploying I2I translation in real-world MRI applications.

\section{Related Works}
Medical I2I translation has been explored using both supervised~\cite{alkan2016magnetic} and self-supervised~\cite{huang2022generalized} learning approaches, with recent advances increasingly favoring generative models. Early studies in generative I2I translation predominantly focused on GANs, which use a generator–discriminator architecture to synthesize realistic images through adversarial training. Within this framework, researchers have tried to enhance translation quality by optimizing input representations and network architectures. For example, ~\cite{kawahara2021t1,vaidya2022perceptually} proposed a conditional GAN for T2w image synthesis from T1w inputs. Specifically, in~\cite{kawahara2021t1} the authors analyze the impact of different input resolutions and dimensionalities. Similarly, Dey \etal~\cite{dey2024mtsr} introduced MTSR-MRI, a GAN-based framework that jointly performs super-resolution and modality translation in a lower-dimensional embedding space.\newline

Diffusion models have emerged as a powerful alternative in generative modeling. Building on the limitations of GANs, these new models provide greater training stability and excel at producing high-fidelity, diverse images by gradually reversing a learned noise process. This advancement has set a new standard for image synthesis, enabling more reliable and realistic results~\cite{dhariwal2021diffusion}. Although these methods have shown success on 2D natural images, a major challenge lies in extending them to the 3D volumetric medical domain while maintaining computational efficiency. Choo \etal~\cite{choo2024slice} tackle this challenge by introducing a slice-consistent 2D diffusion model for CT-to-MRI translation. Their approach incorporates style conditioning and inter-slice trajectory alignment to preserve 3D anatomical coherence, while avoiding direct computation on full 3D volumes. Similarly, MC-IDDPM~\cite{pan2024synthetic} employs a diffusion process for synthetic CT generation from MRI, using a shifted-window transformer V-Net (Swin-VNet) to produce high-quality synthetic CT scans aligned with MRI anatomy.\newline

In recent years, FM models~\cite{lipman2022flow} have been proposed as a generalized framework that enables more efficient training and sampling by learning continuous dynamics to map between source and target data distributions. Several studies have successfully applied flow matching to a range of image-related tasks, including medical image synthesis~\cite{yazdani2025flow}, image restoration~\cite{martin2024pnp}, and natural I2I translation~\cite{chadebec2025lbm}. However, there is limited research specifically focused on applying flow matching to medical I2I translation.

\section{Preliminaries}
We consider the task of medical I2I translation using a dataset of paired T1w and T2w axial slices, denoted as $\{(x^{(i)}, y^{(i)})\}_{i=1}^{N}$, where each $x^{(i)}, y^{(i)} \in \mathbb{R}^{H \times W}$ represents the central axial slice from the T1w and T2w volumes of the $i$-th subject. The objective is to generate the T2w slice $y^{(i)}$ given the corresponding T1w slice $x^{(i)}$, which can be framed as learning a conditional distribution $p(y \mid x = x^{(i)})$. To approximate this distribution, we train a generative model $G_\theta$, parameterized by $\theta$, on the available paired dataset. We evaluate three distinct families of generative models—GANs, diffusion models, and FMs—which we describe in the sections below.

\subsection{Pix2Pix}
The Pix2Pix model~\cite{isola2017image} approaches I2I translation as a supervised learning task using a conditional GAN. The generator $G^{p2p}_\theta$, with parameters $\theta$, learns to translate a T1w slice $x^{(i)}$ into a synthetic T2w slice $\hat{y}^{(i)} = G^{p2p}_\theta(x^{(i)})$, while a discriminator $D_\phi$, with parameters $\phi$, is trained simultaneously to distinguish between real pairs $(x^{(i)}, y^{(i)})$ and fake pairs $(x^{(i)}, \hat{y}^{(i)})$. The training objective combines an adversarial loss with a pixel-wise reconstruction loss:
\begin{equation}
\begin{split}
\mathcal{L}_{p2p} = \mathbb{E}_{x^{(i)}, y^{(i)}} \left[ \log D_\phi(x^{(i)}, y^{(i)}) \right] 
&+ \mathbb{E}_{x^{(i)}} \left[ \log \left(1 - D_\phi(x^{(i)}, G^{p2p}_\theta(x^{(i)}))\right) \right] \\
&+ \lambda \lVert y^{(i)} - G^{p2p}_\theta(x^{(i)}) \rVert_1,
\end{split}
\end{equation}
where $\lambda$ controls the trade-off between realism and fidelity to the ground truth.

\subsection{Conditional Diffusion Model}
Diffusion models~\cite{ho2020denoising} are generative models that learn to reverse a Markovian noising process applied to the data. In our setting, the forward process incrementally corrupts a target T2w slice, $y^{(i)}_0=y^{(i)}$, with Gaussian noise over $T$ steps. At each step $t$, the noisy image $y^{(i)}_t$ is sampled from:

\begin{equation}
    q(y^{(i)}_t \mid y^{(i)}_{t-1}) = \mathcal{N}(y^{(i)}_t; \sqrt{1 - \beta_t}\, y^{(i)}_{t-1}, \beta_t I),
\end{equation}
where $\beta_t$ follows a predefined variance schedule. After $T$ steps, the image is fully transformed into noise, i.e., $y^{(i)}_T \sim \mathcal{N}(0, I)$. The goal of the model is to learn the reverse process that transforms the noise sample $y^{(i)}_T$ back into a realistic T2w image $y^{(i)}_0$. The reverse process $y^{(i)}_t \rightarrow y^{(i)}_{t-1}$ can be expressed by the transition probability: 
\begin{equation}
    p(y^{(i)}_{t-1} \mid y^{(i)}_{t}; \epsilon) = 
\mathcal{N} \left( y^{(i)}_{t-1}; \frac{1}{\sqrt{\alpha_t}} \left(
    y^{(i)}_t - \frac{\beta_t}{\sqrt{1 - \bar{\alpha}_t}}\, \epsilon
    \right), \tilde{\beta}_tI
    \right) 
\end{equation}
where $\alpha_t$, $\bar{\alpha}_t$, and $\tilde{\beta}_t$ are values derived from the variance schedule $\beta_t$ and are defined in~\cite{ho2020denoising}. Here, $\epsilon \sim \mathcal{N}(0, I)$ is the noise used to obtain $y_t^{(i)}$ from the original image $y_0^{(i)}$ during the forward diffusion process. \newline

A neural network $G_\theta(y^{(i)}_t, t)$ is trained to predict the noise $\epsilon$ by minimizing the following loss:

\begin{equation}
    \mathcal{L}_{\epsilon} = \mathbb{E}_{t, y^{(i)}_0, \epsilon \sim \mathcal{N}(0, I)} \left[ \left\| \epsilon - G_\theta(y^{(i)}_t, t) \right\|^2 \right].
\end{equation}

In essence, the model learns to reconstruct a clean image by progressively removing noise, step by step, using a learned denoising function.\newline 

To adapt diffusion models for conditional generation, we incorporate the T1w slice $x^{(i)}$ into the denoising process, and modify the loss term to include the conditioning:

\begin{equation}
    \mathcal{L}_{cdm} = \mathbb{E}_{t, y^{(i)}_0, \epsilon \sim \mathcal{N}(0, I)} \left[ \left\| \epsilon - G^{cdm}_\theta(y^{(i)}_t, t; x^{(i)}) \right\|^2 \right].
\end{equation}

Specifically, we explore two conditioning strategies: (i) concatenating $x^{(i)}$ with the noisy input $y^{(i)}_t$ along the channel dimension at every denoising step, which we refer to as "Concat. Diffusion"; and (ii) employing a ControlNet~\cite{zhang2023adding} module, trained on top of a pre-trained, frozen unconditional diffusion backbone. Specifically, the ControlNet injects conditioning into the decoder layers of the U-Net.

\subsection{Flow Matching}
Given a source (known) distribution $p_0$ and a target (unknown) distribution $p_1$, the goal of FM is to define a time-dependent flow $\psi_t$ that transports a sample $z_0 \sim p_0$ such that $\psi_1(z_0) \sim p_1$. This flow is governed by a time-dependent velocity field $u_t$ through the ordinary differential equation (ODE):
\begin{equation}
\frac{d}{dt} \psi_t(z_0) = u_t(\psi_t(z_0)) \quad
z_0 \sim p_0 \quad
t \in [0,1]
\end{equation}
If $u_t$ is known, the flow $\psi_t$ can be obtained by numerically solving this ODE. In practice, $u_t$ is often unknown or intractable. FM addresses this by learning $u_t$ using a neural network $G_\theta^{fm}$. In its simplest form, FM assumes we have access to a target sample $z_1 \sim p_1$ during training, allowing us to condition the flow entirely on this known endpoint and predetermine where a sample $z_0$ should be transported. This allows us to define a linear path:
\begin{equation}
z_t = \psi_t(z_0) = (1 - t)z_0 + t z_1,
\end{equation}
with a constant, conditional velocity field:
\begin{equation}
u_t(\psi_t(z_0) \mid z_1) = \frac{d z_t}{dt} = z_1 - z_0.
\end{equation}
The model is trained to match the conditional velocity field along this path. This leads to the Conditional Flow Matching (CFM) objective:
\begin{equation}
\mathcal{L}_{cfm} = \mathbb{E}_{t,\, z_0 \sim p_0,\, z_1 \sim p_1} \left[ \left\| G_\theta^{fm}(z_t, t) - (z_1 - z_0) \right\|^2 \right].
\end{equation}
Although this setup requires access to $z_1$ during training, the objective induces the same gradient as the original FM loss involving the true velocity field $u_t$~\cite{lipman2022flow}. As a result, minimizing $\mathcal{L}_{cfm}$ leads to learning correct flow dynamics that generalize to unseen samples at inference time.\newline

We explore two conditioning strategies for employing an FM model to generate a T2w axial slice $y^{(i)}$ from its corresponding T1w slice $x^{(i)}$. The first strategy, which we refer to as \textit{Concat. FM}, initializes the source distribution as $p_0 = \mathcal{N}(0, I)$ and conditions the generation process by concatenating the T1w slice $x^{(i)}$ with the intermediate sample $z_t$ along the channel dimension before regressing the velocity field. The second strategy, which we refer to as \textit{Direct FM}, uses the T1w slice $x^{(i)}$ directly as the source sample $z_0$, learning a mapping to the corresponding T2w slice $y^{(i)}$.

\section{Experimental Design}\label{sec:exp}
This section describes our proposed benchmark study comparing multiple generative models for T1w-to-T2w I2I translation, covering datasets, preprocessing and experimental protocol.

\subsection{Datasets}\label{sec:datasets}
We train and evaluate the models on paired T1w and T2w structural MRI scans from three publicly available datasets: IXI (560 subjects), the Human Connectome Project (HCP, 1002 subjects), and the Cambridge Centre for Ageing and Neuroscience (CamCAN, 633 subjects). All participants across these datasets were healthy adults, with an average age of 41.4±17.7 years and 54\% of subjects being female.

\subsection{Preprocessing}\label{sec:preprocessing}
We process all MRI data through a standardized pipeline designed to ensure consistency, anatomical alignment, and intensity normalization across subjects and imaging modalities. The preprocessing steps include:

\begin{itemize}
    \item \textbf{Bias field correction:} We correct intensity inhomogeneities using the N4ITK algorithm~\cite{tustison2010n4itk}, which mitigates spatially varying signal artifacts and enhances tissue contrast.
    \item \textbf{Brain extraction:} We perform the skull-stripping procedure with SynthStrip~\cite{hoopes2022synthstrip} to isolate brain tissue from non-brain structures, reducing irrelevant variability and focusing subsequent processing on the relevant anatomy.
    \item \textbf{Spatial normalization:} We affinely register each scan to the MNI152 1\,mm$^3$ template using the Advanced Normalization Tools (ANTs)~\cite{avants2008symmetric}, ensuring all images conform to a common anatomical space.
    \item \textbf{Intensity normalization:} We apply WhiteStripe~\cite{shinohara2014statistical} normalization independently to T1w and T2w images, referencing white matter regions to standardize intensity distributions and reduce inter-subject and inter-scan variability.
    \item \textbf{Slice extraction:} For each subject, we extract the central axial slice from both T1w and T2w volumes, and pad it to 224 $\times$ 192 pixels.
\end{itemize}

\subsection{Experimental Protocol}\label{sec:protocol}
We develop and evaluate all generative models within a unified experimental framework to facilitate fair and comprehensive comparison. Key aspects include:

\begin{itemize}
    \item \textbf{Model architecture:} Each model employs a 2D U-Net encoder-decoder backbone with skip connections between corresponding layers. This architecture utilizes two residual blocks at each resolution level, with the number of feature channels progressing from 32, 64, 64, and 64 across its levels. It also integrates attention mechanisms at the latter two resolution levels.
    \item \textbf{Data splits:} We partition the dataset into training (80\%), validation (5\%), and testing (15\%) sets, with strict subject-level separation to prevent data leakage and ensure unbiased evaluation.
    \item \textbf{Training details:} We train all networks with a batch size of 6, selected to optimize GPU memory usage. Training proceeds for up to 300 epochs, with early stopping based on validation loss to prevent overfitting.
    \item \textbf{Evaluation metrics:} We assess model performance using the Structural Similarity Index (SSIM), Mean Squared Error (MSE), and Peak Signal-to-Noise Ratio (PSNR) between the ground-truth and predicted T2w scan. These metrics quantify fidelity, structural preservation, and noise characteristics.
\end{itemize}

\subsection{Implementation Details}
We implement all models using the MONAI framework~\cite{cardoso2022monai}, which facilitates reproducibility. For diffusion-based models, we adopt a standard training configuration with $T = 1000$ denoising steps and a linear variance schedule ranging from $\beta_1 = 1 \times 10^{-4}$ to $\beta_T = 2 \times 10^{-2}$. Ancestral sampling is used during inference. ControlNet is trained and evaluated using the same diffusion schedule and sampling procedure. For the FM-based models, we employ the Euler ODE solver with 300 integration steps. In the Pix2Pix loss, we set $\lambda = 100$. We perform all training and experiments on a Tesla T4 GPU with 16GB of VRAM.

\begin{table}[t]
  \caption{Quantitative results for the T1w-to-T2w I2I translation task. Metrics are reported as mean ± standard deviation across the test set for SSIM, MSE, and PSNR.}
  \label{tab:networkresults}
  \centering
  \def\arraystretch{1.1}
  \setlength{\tabcolsep}{10pt} 
  \begin{tabular}{@{}l|c|c|c@{}}
    \toprule
    \textbf{Method} & \textbf{SSIM} $\uparrow$ & \textbf{MSE} $\downarrow$ & \textbf{PSNR} $\uparrow$ \\
    \hline
    ControlNet~\cite{zhang2023adding} & $0.363 \pm 0.057$ & $0.3887 \pm 0.1427$ & $14.800 \pm 10.498$ \\
    Concat. Diffusion~\cite{ho2020denoising} & $0.469 \pm 0.037$ & $0.5110 \pm 0.2157$ & $15.892 \pm 9.168$ \\
    Concat. FM~\cite{lipman2022flow} & $0.715 \pm 0.007$ & $0.0097 \pm 0.0000$ & $20.326 \pm 1.143$ \\
    Direct FM~\cite{lipman2022flow} & $0.732 \pm 0.001$ & $0.0106 \pm 0.0000$ & $20.011 \pm 1.361$ \\
    Pix2Pix~\cite{isola2017image} & $\mathbf{0.862 \pm 0.001}$ & $\mathbf{0.0054 \pm 0.0000}$ & $\mathbf{22.915 \pm 3.428}$ \\
    \bottomrule
  \end{tabular}
\end{table}

\section{Results}
In our experiments, we conduct three types of analyses: i) a quantitative evaluation focusing on comparing the different approaches in terms of SSIM, MSE, and PSNR performance; ii) a qualitative assessment involving visual comparison of the results; and iii) a comparison of resource allocation with respect to execution time, memory usage, and the number of parameters.

\subsection{Quantitative Comparison}
Table~\ref{tab:networkresults} presents the quantitative results of our benchmark experiment. Pix2Pix demonstrably outperforms all other evaluated methods, achieving the highest SSIM ($0.862$), lowest MSE ($0.0054$), and highest PSNR ($22.915$), collectively indicating superior preservation of structural information, anatomical details, and image quality. FM approaches show moderate but notable performance: Direct FM achieves a slightly higher SSIM ($0.732$) compared to Concat. FM ($0.715$), indicating reasonable reconstruction capabilities for both. However, Concat. FM demonstrates marginally better performance in terms of MSE ($0.0097$ versus $0.0106$). PSNR values are comparable for both variants, with Concat. FM at $20.326$ and Direct FM at $20.011$. Conversely, diffusion-based models show significant performance limitations; the Concat. Diffusion model yields an SSIM of only $0.469$, a substantially elevated MSE of $0.5110$, and a PSNR of $15.892$, indicating considerable challenges in maintaining anatomical fidelity and structural consistency. ControlNet demonstrates the least accurate performance, obtaining the lowest SSIM ($0.363$), the highest MSE ($0.3887$), and the lowest PSNR ($14.800$), suggesting inconsistent output quality and unreliable synthesis capabilities.

\begin{figure}[t]
  \centering
  \includegraphics[width=\linewidth]{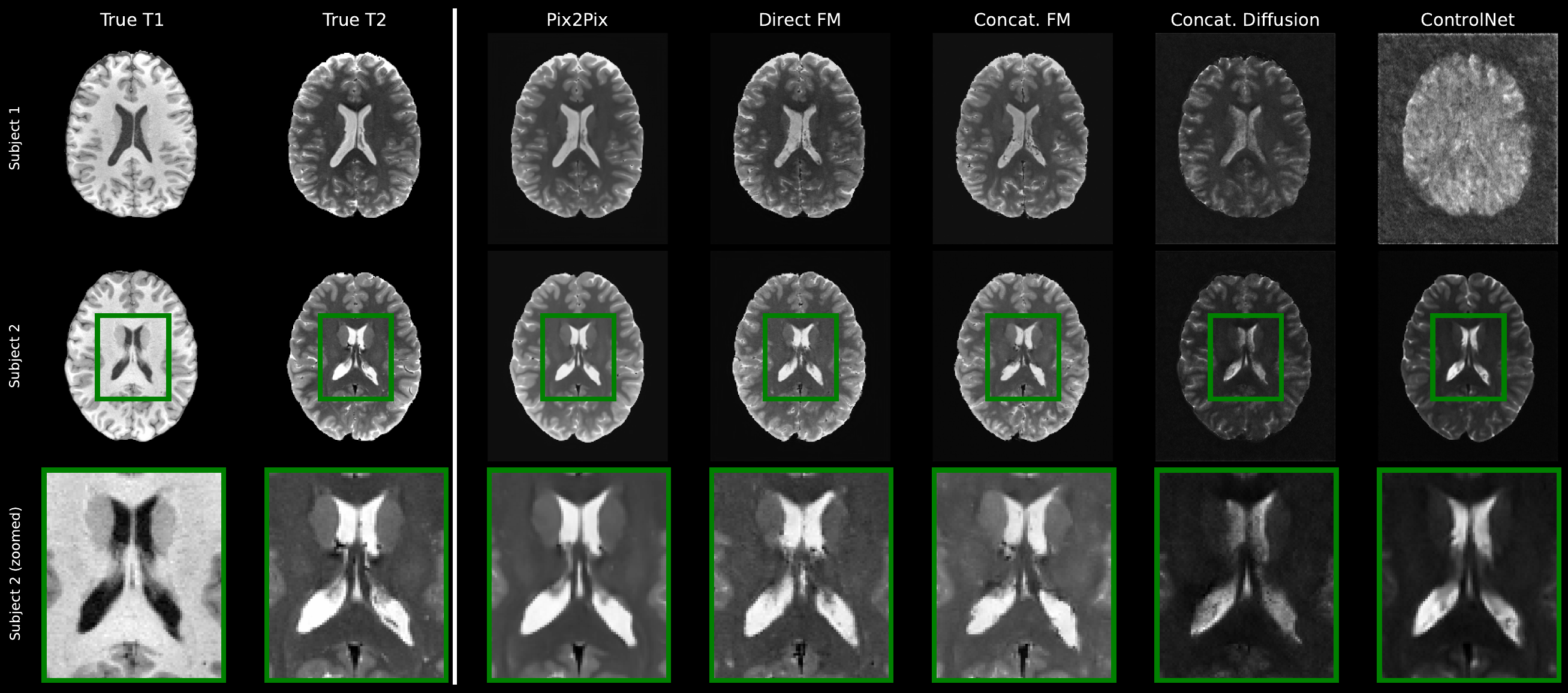}
  \caption{Visual comparison of predictions from different methods across two randomly selected test subjects. The first and second columns display the input T1w and ground-truth T2w slices, respectively. The remaining columns show the predictions from the evaluated generative models. Rows one and two correspond to the two different subjects, while row three displays a zoomed-in region from the second subject, highlighting anatomical details within the brain MRI.}
  \label{fig:non-lesional}
\end{figure}

\begin{figure}[t]
  \centering
  \includegraphics[width=\linewidth]{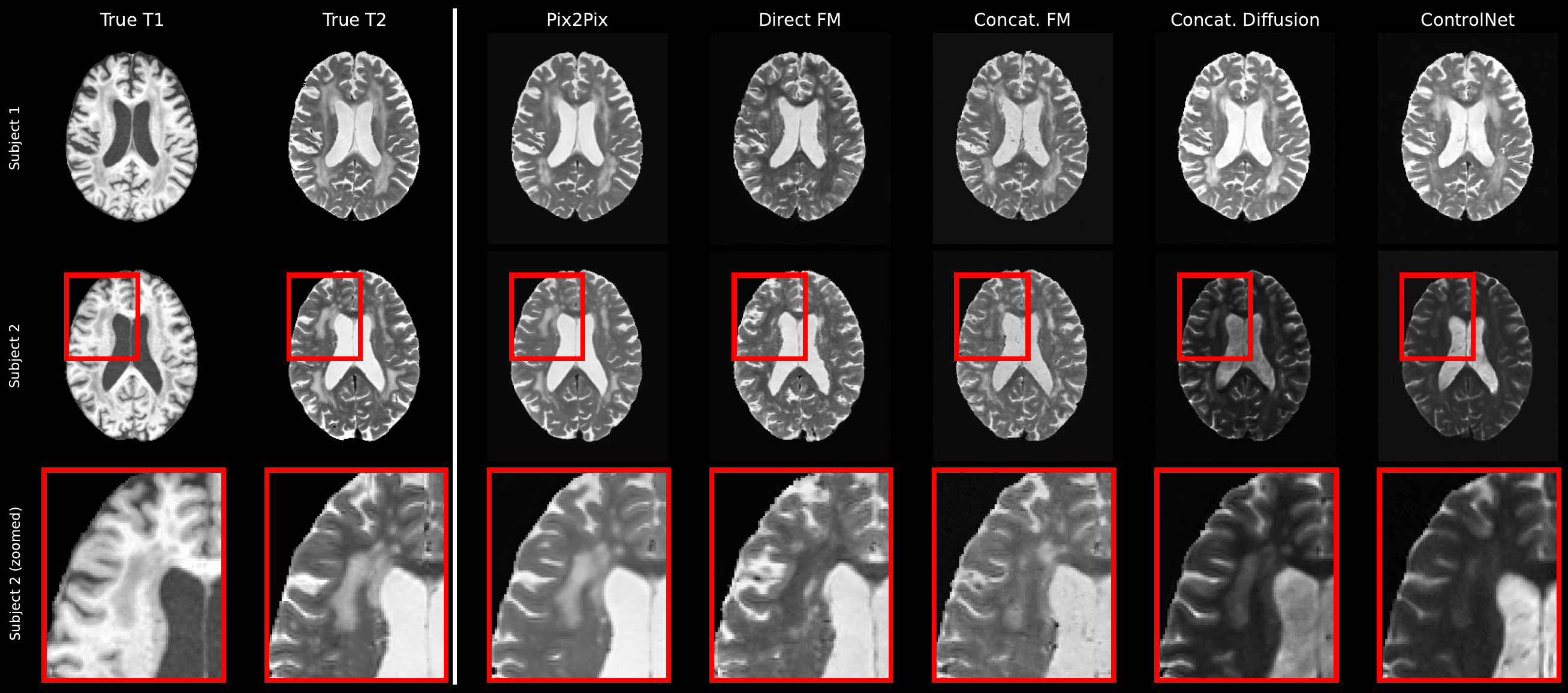}
  \caption{Visual comparison of predictions from different methods across two test subjects that show white matter lesions. The first and second columns display the input T1w and ground-truth T2w slices, respectively. The remaining columns show the predictions from the evaluated generative models. Rows one and two correspond to the two different subjects, while row three displays a zoomed-in region from the second subject, highlighting an area containing white matter lesions.}
  \label{fig:lesional}
\end{figure}

\subsection{Qualitative Assessment and Visual Analysis}
This section provides a visual assessment of the evaluated methods for two case studies, examining (i) the structural fidelity of the predicted output and (ii) the ability to transfer white matter lesions from T1w to T2w scans.

\subsubsection{Evaluating Structural Fidelity}
We visually compare the predicted T2w scans to the ground-truth images in Fig.~\ref{fig:non-lesional}, observing distinct qualitative patterns that confirm our quantitative findings. Pix2Pix consistently produces the most clinically acceptable and realistic synthetic images, demonstrating sharp anatomical boundaries, appropriate tissue contrast differentiation, and minimal noise artifacts. The generated images closely approximate ground truth T2w scans in terms of gray matter–white matter contrast and overall neuroanatomical structure preservation. Flow matching methods demonstrate reasonable structural consistency, though they exhibit slightly increased noise levels and reduced sharpness compared to Pix2Pix. In contrast, diffusion-based models produce images with noticeable artifacts, blurred boundaries, and inconsistent tissue intensities. ControlNet exhibits the most severe qualitative limitations, frequently failing to synthesize coherent T2w images with acceptable anatomical fidelity.

\subsubsection{Evaluating White Matter Lesions Transfer}
The evaluation of subjects with pronounced white matter lesions, as shown in Figure~\ref{fig:lesional}, yields critical insights into each method's capability to transfer pathological features. Pix2Pix maintains superior performance, accurately translating lesion visibility and contrast characteristics. FM methods, notably Direct FM, sometimes perform lesion inpainting rather than faithful reconstruction, indicating a potential misinterpretation of lesions as artifacts. Both diffusion-based models consistently fail to accurately represent pathological features, as they are not able to transfer the lesion accurately.

\begin{table}[t]
  \caption{Computational efficiency comparison of generative models for the T1w-to-T2w I2I task. Metrics reported include the number of parameters, inference time (Inf. Time), and inference memory usage (Inf. Memory).}
  \label{tab:computationalresults}
  \centering
  \def\arraystretch{1.1}
  \setlength{\tabcolsep}{10pt} 
  \begin{tabular}{@{}l|c|c|c@{}}
    \toprule
    \textbf{Method} & \textbf{\#Params.} $\downarrow$ & \textbf{Inf. Time} $\downarrow$ & \textbf{Inf. Memory} $\downarrow$ \\
    \hline
    ControlNet~\cite{zhang2023adding} & 3,255,745 & 57.87 s & 12.42 MB \\
    Concat. Diffusion~\cite{ho2020denoising} & 2,328,737 & 41.33 s & 8.88 MB \\
    Concat. FM~\cite{lipman2022flow} & 2,328,737 & 11.94 s & 8.88 MB \\
    Direct FM~\cite{lipman2022flow} & 2,328,449 & 11.61 s & 8.88 MB \\
    Pix2Pix~\cite{isola2017image} & 2,328,449 & 0.05 s & 8.88 MB \\
    \bottomrule
  \end{tabular}
\end{table}

\subsection{Resources Allocation at Inference Time}
Table~\ref{tab:computationalresults} presents a comparative overview of the computational resources required by each generative model at inference time, including the number of parameters, computational time, and memory usage. Among the evaluated methods, Pix2Pix is the most resource-efficient, requiring only 2,328,449 parameters for the generator network, an extremely fast inference time of 0.05 seconds, and a low memory footprint of 8.88 MB. In contrast, ControlNet and the standard diffusion model are notably less efficient. ControlNet demands the highest number of parameters (3,255,745), the longest inference time (57.87 seconds), and the largest memory usage (12.42 MB). Concat Diffusion, while slightly more efficient than ControlNet, still lags behind both the flow matching and GAN-based approaches. Both flow matching variants (Concat. FM and Direct FM) offer a balanced compromise. They match Pix2Pix in memory usage and parameter count, but have inference times (around 11–12 seconds) that are significantly longer than Pix2Pix, though much faster than diffusion and ControlNet. Overall, Pix2Pix is clearly the best method in terms of computational efficiency, combining the fastest inference, lowest memory use, and smallest parameter count. This makes it highly suitable for real-time or resource-constrained clinical applications.

\section{Conclusion}
The synthesis of T2w MRI from T1w scans marks a significant step forward in medical imaging. This approach addresses key clinical challenges by reducing scan times, lowering costs, and improving access to comprehensive imaging. T1w images are best for anatomical detail, while T2w images excel at identifying fluid and pathology. Reliable cross-modal synthesis could decrease the need for multiple MRI acquisitions while maintaining diagnostic quality.\newline

Our benchmarking results show that Pix2Pix, a GAN-based approach, consistently outperforms both FM and diffusion-based methods across all evaluated metrics. Although prior work has reported that flow-based models can surpass GANs when trained on large-scale datasets~\cite{dhariwal2021diffusion}, recent studies~\cite{bertrand2025closed,akbar2025beware} have highlighted their tendency to overfit and memorize the training distribution under low-data regimes or simplified tasks. In our setting which is characterized by relatively small datasets and low-dimensional 2D axial slices, this overfitting behavior was particularly pronounced in flow-based models. Therefore, our findings may not generalize to other settings involving larger datasets, higher-dimensional data, or more complex tasks.\newline

For this reason, future work could investigate model performance across varying dataset sizes and a range of data dimensionalities—from low-resolution to high-resolution 3D MRI—as well as evaluate the generalizability of these models to out-of-distribution data. Finally, expanding the scope of this work by including other modalities (\eg, CT, PET) will aid in developing future methods and broaden the applicability of I2I frameworks. Such extensions could open new avenues for clinical and research use.

\subsubsection{\ackname}
The work of Francesco Guarnera has been supported by MUR in the framework of PNRR PE0000013, under project “Future Artificial Intelligence Research – FAIR”. This work is partially supported by the Horizon Europe ``Open source deep learning platform dedicated to Embedded hardware and Europe'' project (Grant Agreement, project 101112268 - NEUROKIT2E)

%
%
\bibliographystyle{splncs04}
\bibliography{main}

\begin{thebibliography}{10}
\providecommand{\url}[1]{\texttt{#1}}
\providecommand{\urlprefix}{URL }
\providecommand{\doi}[1]{https://doi.org/#1}

\bibitem{akbar2025beware}
Akbar, M.U., Wang, W., Eklund, A.: Beware of diffusion models for synthesizing medical images—a comparison with gans in terms of memorizing brain mri and chest x-ray images. Machine Learning: Science and Technology  \textbf{6}(1),  015022 (2025)

\bibitem{alkan2016magnetic}
Alkan, C., Cocjin, J., Weitz, A.: Magnetic resonance contrast prediction using deep learning. Google Scholar  (2016)

\bibitem{avants2008symmetric}
Avants, B.B., Epstein, C.L., Grossman, M., Gee, J.C.: Symmetric diffeomorphic image registration with cross-correlation: evaluating automated labeling of elderly and neurodegenerative brain. Medical image analysis  \textbf{12}(1),  26--41 (2008)

\bibitem{bertrand2025closed}
Bertrand, Q., Gagneux, A., Massias, M., Emonet, R.: On the closed-form of flow matching: Generalization does not arise from target stochasticity. arXiv preprint arXiv:2506.03719  (2025)

\bibitem{cardoso2022monai}
Cardoso, M.J., Li, W., Brown, R., Ma, N., Kerfoot, E., Wang, Y., Murrey, B., Myronenko, A., Zhao, C., Yang, D., et~al.: Monai: An open-source framework for deep learning in healthcare. arXiv preprint arXiv:2211.02701  (2022)

\bibitem{chadebec2025lbm}
Chadebec, C., Tasar, O., Sreetharan, S., Aubin, B.: Lbm: Latent bridge matching for fast image-to-image translation. arXiv preprint arXiv:2503.07535  (2025)

\bibitem{choo2024slice}
Choo, K., Jun, Y., Yun, M., Hwang, S.J.: Slice-consistent 3d volumetric brain ct-to-mri translation with 2d brownian bridge diffusion model. In: International Conference on Medical Image Computing and Computer-Assisted Intervention. pp. 657--667. Springer (2024)

\bibitem{dey2024mtsr}
Dey, A., Ebrahimi, M.: Mtsr-mri: Combined modality translation and super-resolution of magnetic resonance images. In: Medical Imaging with Deep Learning. pp. 743--757. PMLR (2024)

\bibitem{dhariwal2021diffusion}
Dhariwal, P., Nichol, A.: Diffusion models beat gans on image synthesis. Advances in neural information processing systems  \textbf{34},  8780--8794 (2021)

\bibitem{ho2020denoising}
Ho, J., Jain, A., Abbeel, P.: Denoising diffusion probabilistic models. Advances in neural information processing systems  \textbf{33},  6840--6851 (2020)

\bibitem{hoopes2022synthstrip}
Hoopes, A., Mora, J.S., Dalca, A.V., Fischl, B., Hoffmann, M.: Synthstrip: skull-stripping for any brain image. NeuroImage  \textbf{260},  119474 (2022)

\bibitem{huang2022generalized}
Huang, Y., Zheng, F., Sun, X., Li, Y., Shao, L., Zheng, Y.: Generalized brain image synthesis with transferable convolutional sparse coding networks. In: {ECCV}. pp. 183--199. Springer (2022)

\bibitem{isola2017image}
Isola, P., Zhu, J.Y., Zhou, T., Efros, A.A.: Image-to-image translation with conditional adversarial networks. In: Proceedings of the IEEE conference on computer vision and pattern recognition. pp. 1125--1134 (2017)

\bibitem{kawahara2021t1}
Kawahara, D., Nagata, Y.: T1-weighted and t2-weighted mri image synthesis with convolutional generative adversarial networks. reports of practical Oncology and radiotherapy  \textbf{26}(1),  35--42 (2021)

\bibitem{lipman2022flow}
Lipman, Y., Chen, R.T., Ben-Hamu, H., Nickel, M., Le, M.: Flow matching for generative modeling. arXiv preprint arXiv:2210.02747  (2022)

\bibitem{martin2024pnp}
Martin, S., Gagneux, A., Hagemann, P., Steidl, G.: Pnp-flow: Plug-and-play image restoration with flow matching. arXiv preprint arXiv:2410.02423  (2024)

\bibitem{pan2024synthetic}
Pan, S., Abouei, E., Wynne, J., Chang, C.W., Wang, T., Qiu, R.L., Li, Y., Peng, J., Roper, J., Patel, P., et~al.: Synthetic ct generation from mri using 3d transformer-based denoising diffusion model. Medical Physics  \textbf{51}(4),  2538--2548 (2024)

\bibitem{ronneberger2015u}
Ronneberger, O., Fischer, P., Brox, T.: U-net: Convolutional networks for biomedical image segmentation. In: Medical image computing and computer-assisted intervention--MICCAI 2015: 18th international conference, Munich, Germany, October 5-9, 2015, proceedings, part III 18. pp. 234--241. Springer (2015)

\bibitem{shinohara2014statistical}
Shinohara, R.T., Sweeney, E.M., Goldsmith, J., Shiee, N., Mateen, F.J., Calabresi, P.A., Jarso, S., Pham, D.L., Reich, D.S., Crainiceanu, C.M., et~al.: Statistical normalization techniques for magnetic resonance imaging. NeuroImage: Clinical  \textbf{6},  9--19 (2014)

\bibitem{tustison2010n4itk}
Tustison, N.J., Avants, B.B., Cook, P.A., Zheng, Y., Egan, A., Yushkevich, P.A., Gee, J.C.: N4itk: improved n3 bias correction. IEEE transactions on medical imaging  \textbf{29}(6),  1310--1320 (2010)

\bibitem{vaidya2022perceptually}
Vaidya, A., Stough, J.V., Patel, A.A.: Perceptually improved t1-t2 mri translations using conditional generative adversarial networks. In: Medical Imaging 2022: Image Processing. vol. 12032, pp. 505--511. SPIE (2022)

\bibitem{yazdani2025flow}
Yazdani, M., Medghalchi, Y., Ashrafian, P., Hacihaliloglu, I., Shahriari, D.: Flow matching for medical image synthesis: Bridging the gap between speed and quality. arXiv preprint arXiv:2503.00266  (2025)

\bibitem{zhang2023adding}
Zhang, L., Rao, A., Agrawala, M.: Adding conditional control to text-to-image diffusion models. In: Proceedings of the IEEE/CVF international conference on computer vision. pp. 3836--3847 (2023)

\end{thebibliography}
\end{document}